\PassOptionsToPackage{dvipsnames}{xcolor}  
\documentclass[conference]{ecai}

\usepackage{latexsym}
\usepackage{amssymb}
\usepackage{amsmath}
\usepackage{amsthm}
\usepackage{enumitem}
\usepackage{graphicx}
\usepackage{color}
\usepackage{algorithm}

\usepackage{algorithmic}

\usepackage{float}
\usepackage{tabularx}
\usepackage{booktabs}
\usepackage{afterpage}
\usepackage{pgfplots}
\usepackage{pgfplotstable}
\pgfplotsset{compat=1.17}

\usepackage[nolist,nohyperlinks]{acronym}
\usepackage{multirow}
\usepackage{xargs}                      

\usepackage[colorinlistoftodos,prependcaption,textsize=tiny]{todonotes}
\newcommandx{\unsure}[2][1=]{\todo[linecolor=red,backgroundcolor=red!25,bordercolor=red,#1]{#2}}
\newcommandx{\change}[2][1=]{\todo[linecolor=blue,backgroundcolor=blue!25,bordercolor=blue,#1]{#2}}
\newcommandx{\info}[2][1=]{\todo[linecolor=OliveGreen,backgroundcolor=OliveGreen!25,bordercolor=OliveGreen,#1]{#2}}
\newcommandx{\improvement}[2][1=]{\todo[linecolor=Plum,backgroundcolor=Plum!25,bordercolor=Plum,#1]{#2}}
\newcommandx{\thiswillnotshow}[2][1=]{\todo[disable,#1]{#2}}

\DeclareUnicodeCharacter{202F}{\,}  
\DeclareUnicodeCharacter{2013}{--}  
\DeclareUnicodeCharacter{2014}{---} 
\DeclareUnicodeCharacter{00D7}{\times}
\DeclareUnicodeCharacter{2032}{'}
\DeclareUnicodeCharacter{2026}{\ldots}

\acrodef{ML}{Machine Learning}
\acrodef{dToU}{dynamic Time-of-Use}
\acrodef{GAN}{Generative Adversarial Networks}
\acrodef{WGAN}{Wasserstein–GP Generative Adversarial Networks}
\acrodefplural{GAN}[GANs]{Generative Adversarial Networks}
\acrodef{MIA}{Membership Inference Attacks}
\acrodef{D}{Discriminator}
\acrodef{G}{Generator}
\acrodef{SVR}{Support Vector Regression}
\acrodef{SVM}{Support Vector Machine}
\acrodef{RMSE}{Rooted Mean Squared Error}
\acrodefplural{SVM}[SVMs]{Support Vector Machine}
\acrodef{CTGAN}{Conditional Tabular GAN}

\newcommand{\BibTeX}{B\kern-.05em{\sc i\kern-.025em b}\kern-.08em\TeX}

\begin{document}


\begin{frontmatter}


\paperid{1756} 


\title{Evaluating Privacy-Utility Tradeoffs in Synthetic Smart Grid Data\\
}


\author[A]{\fnms{Andre}~\snm{Catarino}\thanks{Corresponding Author. Email: up202408593@up.pt}}
\author[A,B,C]{\fnms{Rui}~\snm{Melo}\thanks{Work done as an INESC-ID external collaborator.}}
\author[A,B]{\fnms{Rui}~\snm{Abreu}} 
\author[D]{\fnms{Luis}~\snm{Cruz}} 

\address[A]{Faculty of Engineering, University of Porto}
\address[B]{INESC-ID}
\address[C]{Center for Responsible AI}
\address[D]{TU Delft}


\begin{abstract}
The widespread adoption of \ac{dToU} electricity tariffs requires accurately identifying households that would benefit from such pricing structures. However, the use of real consumption data poses serious privacy concerns, motivating the adoption of synthetic alternatives. In this study, we conduct a comparative evaluation of four synthetic data generation methods, \ac{WGAN}, \ac{CTGAN}, Diffusion Models, and Gaussian noise augmentation, under different synthetic regimes. We assess classification utility, distribution fidelity, and privacy leakage. Our results show that architectural design plays a key role: diffusion models achieve the highest utility (macro-F1 up to 88.2\%), while \ac{CTGAN} provide the strongest resistance to reconstruction attacks. These findings highlight the potential of structured generative models for developing privacy-preserving, data-driven energy systems.
\end{abstract}

\end{frontmatter}


\section{Introduction}

The transition toward sustainable energy systems necessitates effective demand-side management strategies. One such approach is the adoption of dynamic electricity tariffs, including \ac{dToU} tariffs, which incentivise consumers to shift electricity usage from peak to off-peak periods. This not only improves grid efficiency and facilitates renewable energy integration but also helps reduce carbon emissions. However, deploying such demand-side strategies relies heavily on detailed household consumption data, typically collected via smart meters, raising significant privacy concerns.

Synthetic data generation has emerged as a promising solution to balance the need for detailed data analytics with privacy preservation. In this context, generative models offer the ability to create realistic but artificial energy consumption data, enabling utility providers and researchers to conduct meaningful analyses without exposing sensitive user information. \acp{WGAN}~\cite{goodfellow2020generative}, \ac{CTGAN}~\cite{mirza2014conditional}, Diffusion Models~\cite{sohldickstein2015deepunsupervisedlearningusing}, and Gaussian Noise Augmentation~\cite{sensor_data_augmentation} represent a spectrum of approaches for synthetic data generation, each with distinct advantages for modelling temporal and structured data.

While synthetic data techniques aim to protect user privacy, they are not immune to privacy attacks. \ac{MIA}~\cite{shokri2017membership} attempt to identify whether specific data points were part of a model’s training set, while reconstruction attacks aim to recover sensitive information from model outputs. Therefore, rigorous evaluation of these vulnerabilities is essential for assessing the trade-off between data utility and privacy.

Our work extends the application of synthetic data augmentation from load forecasting to the prediction of household suitability for dynamic tariffs, an area less explored in current literature~\cite{ctgan_short_term_forecasting}. We systematically benchmark multiple generative models with respect to their impact on classification accuracy and resilience to privacy attacks. To our knowledge, this is the first study to holistically evaluate the privacy-utility trade-offs of synthetic data in the context of tariff suitability rather than conventional load prediction. Our contributions aim to inform data-driven energy policy and promote privacy-aware research in smart grid environments.

To guide our investigation, we formulate the following:
\begin{itemize}
    \item \textbf{RQ1:} Can machine learning models accurately classify households based on their suitability for \ac{dToU} tariffs using behavioural features?
    \item \textbf{RQ2:} What is the impact of synthetic data augmentation on the predictive performance of tariff classification models?
    \item \textbf{RQ3:} How effective are synthetic data generation strategies under privacy attacks?
\end{itemize}

\section{Related Work}
Victor von Loessl~\cite{privacy_concerns} analysed survey data from German households, revealing that only 23.6\% of respondents exhibited low privacy concerns, primarily due to transparent data-handling practices. These practices help mitigate aversion to smart meter-enabled tariffs, particularly among consumers with heightened privacy concerns.

The effectiveness of dynamic tariffs in influencing household energy consumption depends heavily on tariff design. Freier et al.~\cite{dynamic_tariffs_design} proposed a methodology for structuring time-varying electricity prices to balance supply and demand while promoting renewable energy integration. Their findings indicated that short-term price variations play a crucial role in financial savings, yet households with limited flexibility struggle to achieve significant benefits. Similarly, researchers in \cite{dynamic_tariffs_response} developed a two-stage game-theoretic model demonstrating that dynamic tariffs enhance market efficiency by aligning retail and wholesale prices. However, they highlight the need for regulatory interventions to ensure fair redistribution of economic gains among consumers.

Regarding privacy, concerns over smart meter data have driven the development of synthetic data generation techniques. Sheng Chai et al.~\cite{chai2024faradaysyntheticsmartmeter} introduced Faraday, a Variational Autoencoder (VAE)-based model trained on 300 million UK smart meter readings. This model generates synthetic load profiles that closely resemble real energy consumption patterns while ensuring user privacy. Additionally, recent advances in generative models have further improved synthetic data quality for smart grid applications. Bilgi Yilmaz et al.~\cite{synthetic_demand_gan} explored RCGAN, TimeGAN, CWGAN, and RCWGAN

In contrast, work by Asma Maalejand and Chiheb Rebai~\cite{sensor_data_augmentation} introduced a Gaussian noise-based augmentation strategy for \ac{SVR} models, demonstrating the utility of noise-based techniques to improve the performance of machine learning models in energy forecasting tasks.



\section{Methodology}

\begin{figure}[htbp]
\centering
\includegraphics[height=5.3cm]{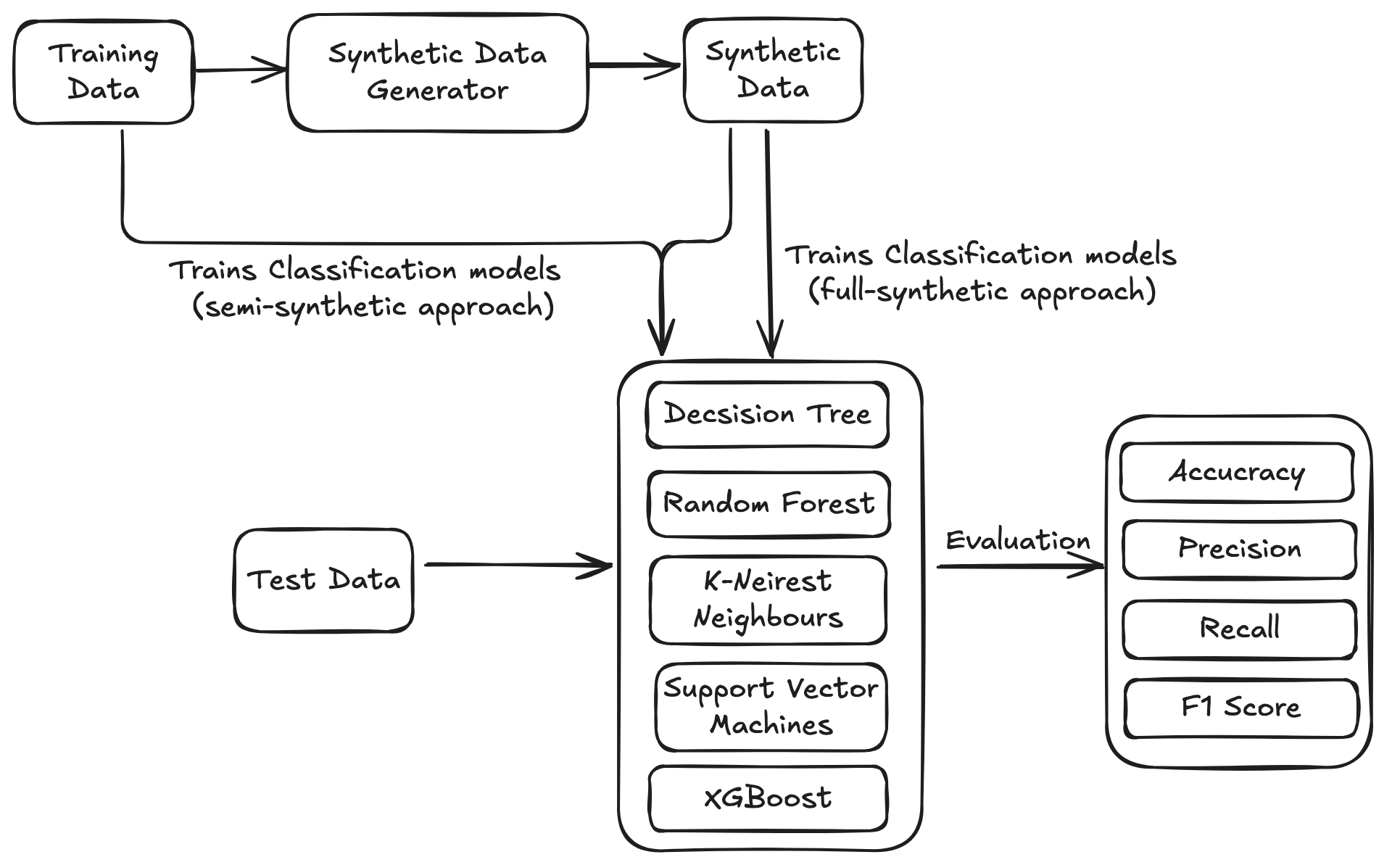}
\caption{Proposed methodology overview.}
\label{fig:methodology}
\end{figure}

Our approach follows a structured pipeline designed to ensure accurate and privacy-preserving predictions of household suitability for \ac{dToU} tariffs. The methodology consists of multiple stages, including data preprocessing, feature engineering, the experimental details of the synthetic data generation and model training, and the evaluation framework, as highlighted in Figure ~\ref{fig:methodology}.

\subsection{Data Preprocessing} 
The dataset originates from the UK Power Networks' Low Carbon London project (2011–2014), comprising 167 million half-hourly electricity consumption records from 5,567 households in Greater London. It includes two groups: approximately 1,100 households enrolled in a \ac{dToU} tariff trial in 2013, receiving real-time price signals, and around 4,500 households on a fixed-rate tariff serving as a control group. The \ac{dToU} tariff featured variable pricing tiers, influencing household consumption patterns based on peak and off-peak rates. To enable meaningful comparisons across households with varying energy consumption levels, we standardised each consumption value.

\subsection{Problem Formulation and Label Construction}

Let each household \( h_i \in \mathcal{H} \) be described by a vector of behavioural features \( \mathbf{x}_i = [x_i^{(1)}, x_i^{(2)}, \ldots, x_i^{(d)}] \in \mathbb{R}^d \), derived from smart meter readings under dynamic pricing conditions. These include metrics such as the ratio of energy consumed during high-tariff periods, load entropy, and changes in weekday/weekend usage. The goal is to learn a binary function \( f: \mathbb{R}^d \rightarrow \{0,1\} \), where \( f(\mathbf{x}_i) = 1 \) denotes that the household is \emph{responsive} (i.e., likely to shift consumption in response to price signals), and \( f(\mathbf{x}_i) = 0 \) denotes otherwise.

However, the real data \( \mathcal{D}_{\text{real}} = \{(\mathbf{x}_i, y_i)\} \) poses privacy risks due to its granularity. To mitigate this, we introduce a synthetic dataset \( \mathcal{D}_{\text{syn}} = \{(\tilde{\mathbf{x}}_j, \tilde{y}_j)\} \), generated via models such as \ac{WGAN}, Diffusion, and noise-based augmentations. 

Since explicit ground truth labels for responsiveness are unavailable, we adopt an unsupervised scoring method. First, all behavioural features are standardised using z-score normalisation. We then apply Principal Component Analysis (PCA) and extract the first principal component (PC1), which captures the dominant axis of variance across households. This yields a scalar \emph{responsiveness score} for each household:

\begin{equation}
s_i = \sum_{j=1}^d \mathbf{w_j} \cdot z_i^{(j)}
\end{equation}

where \( z_i^{(j)} \) is the standardized value of feature \( j \) for household \( i \), and \( \mathbf{w_j} \) is the loading (coefficient) of that feature in PC1. Households with higher \( s_i \) values exhibit behaviours more aligned with the dominant pattern of tariff responsiveness.

To obtain binary labels, we set a threshold \( q \) corresponding to the 75th percentile of \( s_i \) values across all households:

\begin{equation}
\text{Responsive}_i = \mathbb{I}[s_i > \text{Quantile}_{q}(s)]
\end{equation}

where \( \mathbb{I}[\cdot] \) is the indicator function. Households with scores above this threshold are labelled as responsive (\(1\)); others are labelled as non-responsive (\(0\)). In our experiments, we set \( q = 0.75 \), capturing the top 25\% of households with the most favourable behavioural indicators.

While ground truth responsiveness labels are not available, our use of PCA over engineered behavioural features is consistent with prior work in behavioural segmentation and responsiveness inference in smart grid contexts. Similar dimensionality-reduction and clustering techniques have been used to derive latent consumption patterns or segment user behaviour in demand response studies~\cite{BECKEL2014397, 6545387}. The resulting scores align well with expert intuition and tariff policy objectives, offering a practical proxy for real-world suitability.

Features included in the construction of the composite responsiveness score are the following:

\begin{itemize}
    \item \texttt{high\_usage\_ratio} – fraction of total energy used during high-tariff periods; penalized in the score.
    \item \texttt{low\_usage\_ratio} – proportion of consumption during low-tariff periods; rewarded in the score.
    \item \texttt{peak\_hour\_ratio} – share of usage during peak daily hours (16:00–20:00); lowers the responsiveness score.
    \item \texttt{weekend\_shift} – difference in average consumption between weekends and weekdays; large shifts reduce score due to behavioural inconsistency.
    \item \texttt{load\_entropy} – entropy of usage distribution; moderate entropy indicates more regular, responsive behaviour.
    \item \texttt{load\_factor\_low} – efficiency of consumption during low-tariff periods; higher values are rewarded.
\end{itemize}

Households that exhibit reduced electricity usage during high-tariff periods and demonstrate more regular load patterns, such as a higher load factor during low-tariff periods, receive higher responsiveness weights. Positive PC1 loadings (e.g., \texttt{low\_usage\_ratio}, \texttt{load\_factor\_low}) contribute positively to the responsiveness score, indicating alignment with \ac{dToU} incentives. In contrast, negative loadings (e.g., \texttt{high\_usage\_ratio}, \texttt{peak\_hour\_ratio}) reflect behaviours that are less responsive to such incentives.

\subsection{Autocorrelation}
\begin{figure}[h]
\centering
\includegraphics[width=0.45\textwidth]{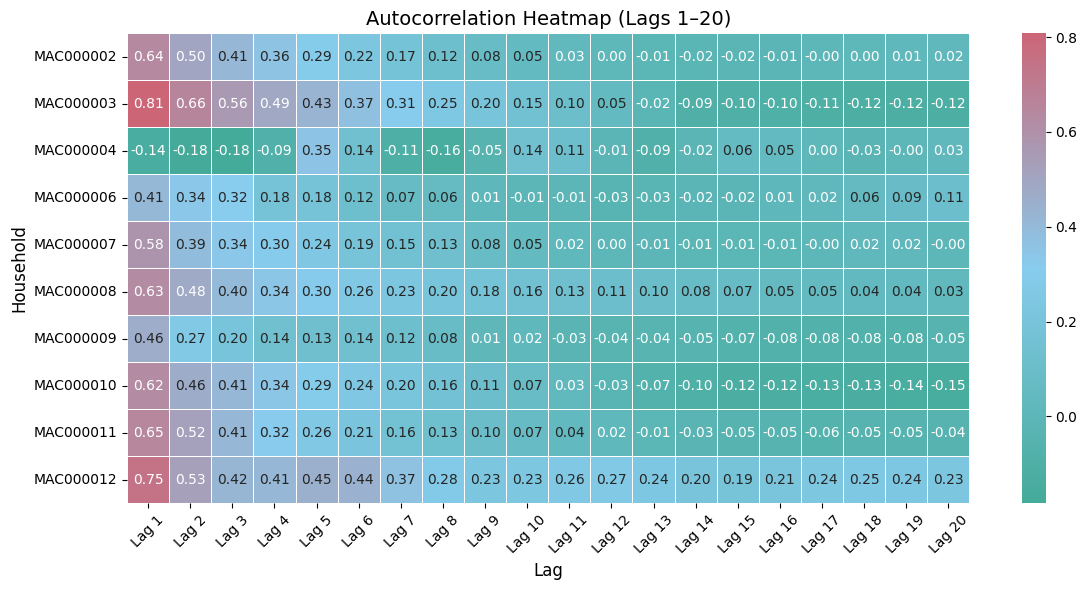}
\caption{Consumption autocorrelation heatmap for a household subsample up to lag 20. Warmer regions indicate stronger correlations, reflecting patterns in energy usage over time.}
\label{fig:autocorrelation_heatmap}
\end{figure}

Autocorrelation measures the relationship between a household's past and present energy consumption, revealing recurring patterns and temporal dependencies. Key features include \textbf{Mean Autocorrelation}, which captures overall consumption stability, \textbf{Max Autocorrelation}, indicating the strongest periodicity, and \textbf{Decay Rate}, reflecting how quickly past usage loses influence. To illustrate the strength of temporal dependencies, Figure~\ref{fig:autocorrelation_heatmap} presents a heatmap of consumption autocorrelation for a household subset.

\subsubsection{Baseline Classifiers and Evaluation Protocol}
\label{sec:eval:baselines}

We benchmark five standard tabular classifiers: Decision Tree (DT), Random Forest (RF), $k$‑Nearest Neighbours (KNN), Support‑Vector Machine with an \emph{RBF} kernel (\acs{SVM}), and Extreme Gradient Boosting (\textsc{XGBoost}), using a fixed nested‑cross‑validation pipeline.
Each estimator is wrapped in a \textsc{StandardScaler} and tuned by \textbf{randomised search} (\texttt{n\_iter}$=10$) on an inner \emph{3‑fold} stratified CV.  
Performance is assessed on an outer \textbf{5‑fold} stratified CV.

\paragraph{Metric.}  
Our primary utility measure is the \textbf{macro‑F1 score}

\begin{equation}
\mathrm{F1}_{\text{macro}} \;=\; \frac{1}{C} \sum_{c=1}^{C} \frac{2\,\mathrm{Precision}_{c}\,\mathrm{Recall}_{c}}{\mathrm{Precision}_{c}+\mathrm{Recall}_{c}}
\end{equation}

with \(C=2\) classes.  Macro‑F1 weights all classes equally, making it the appropriate choice for our moderately imbalanced dataset (class ratio 3:1).  
For each classifier, we report $\mu_{\mathrm{F1}} \pm \sigma$ and the 95\% confidence interval across the five outer folds.

\vspace{2pt}
Hyperparameter search spaces appear in Table~\ref{tab:grids}.  Code and data are publicly available at our repository\footnote{Omitted for review.}.


\subsection{Generative Data Synthesis Approaches}
\label{sec:exp:generators}

\textbf{Notation.} We denote the numerical feature matrix by $X\in\mathbb{R}^{n\times d}$ and the binary target by $y\in\{0,1\}^{n}$.  A generator produces $(\tilde{X},\tilde{y})$ of the same dimensionality.

\begin{itemize}[leftmargin=1.1em]
\label{sec:exp:wgan}
    \item \textbf{(1) Wasserstein–GP GAN} Our \ac{WGAN} comprises a generator~$G$ and a critic~$C$ (\emph{three‑layer} MLPs, hidden 128, \textsc{ReLU}).  Spectral norm imposes the 1‑Lipschitz constraint on $C$, and we train with the Wasserstein loss plus gradient penalty ($\lambda_{\mathrm{gp}}=10$)~\cite{gulrajani2017improvedtrainingwassersteingans}.
    To avoid label‑mode collapse, we add:
    \begin{enumerate}
      \item Entropy maximisation, incorporating an entropy regularisation term into the generator's loss function, inspired by methods that maximise the entropy of generated distributions to enhance sample diversity ~\cite{khorramshahi2020gansvariationalentropyregularizers}.
      \item Label‑balance term $\bigl(\operatorname{E}_{\tilde{y}}[\tilde{y}]-p_{r}\bigr)^{2}$,
      \item MSE to the empirical class ratio $p_{r}$.
    \end{enumerate}
    
    We run five critical updates per generator update with Adam optimiser~\cite{kingma2017adammethodstochasticoptimization} ($\alpha=10^{-4},\;\beta_{1}=0,\;\beta_{2}=0.9$), batch size 32 for 100 epochs. Preliminary ablation experiments revealed that removing the entropy and class-balance regularizers led to severe mode collapse, particularly in the label distribution. In early trials, the generator defaulted to producing only the majority class. These regularizers were therefore retained in all \ac{WGAN} experiments to ensure balanced class synthesis and stable training.

    \item \textbf{(2) DDPM-Based Diffusion Model} Our diffusion-based generator is inspired by the Denoising Diffusion Probabilistic Model (DDPM) \cite{ho2020denoisingdiffusionprobabilisticmodels} framework and adapted for the tabular classification setting. The model employs a three-layer MLP backbone (hidden size 128) with ReLU activations, and incorporates time-step conditioning via a learned embedding layer ($t \in \{0, \dots, T{-}1\}$, with $T{=}100$ steps). The architecture jointly predicts two outputs: (i) the additive Gaussian noise $\varepsilon_{\theta}(x_t, t)$ and (ii) the binary class logit $\hat{y}$, using a dual-head design.
    
    We train the model using a composite objective:
    \begin{equation}
      \mathcal{L} = \text{MSE}(\varepsilon, \hat{\varepsilon}) + \text{BCE}(y, \hat{y})        
    \end{equation}

    balancing both the denoising accuracy and classification capability. The optimiser is Adam ($\alpha{=}10^{-3}$, batch size 32), with a training horizon of 50 epochs.
    
    To improve generation stability, we maintain an Exponential Moving Average (EMA) of model weights (decay 0.999), applied during inference. Sampling follows a deterministic DDIM-style reverse process. Starting from Gaussian noise $x_T \sim \mathcal{N}(0, I)$, the model refines the sample through time-stepped denoising conditioned on the learned time embeddings. Generated features are decoded through the learned denoising path, and class labels are derived from the class logit head using a thresholded sigmoid output.
    
    This formulation allows for joint synthetic feature-label generation in a time-consistent, noise-aware manner, while incorporating practical design elements like EMA smoothing and dual-task training to ensure stable and realistic synthetic data production.

    \item \textbf{(3) \ac{CTGAN}}
\label{sec:exp:ctgan}

We adopt the \textsc{SDV} \texttt{CTGANSynthesizer} implementation of \ac{CTGAN} \cite{xu2019modeling}, designed to handle the unique challenges of mixed-type tabular data. CTGAN models both continuous and categorical features by conditioning the generator and discriminator on sampled values of discrete columns, allowing for realistic class‑conditional generation.

We specify metadata using the \texttt{SingleTableMetadata} interface, registering all continuous behavioural features with `numerical` type and the binary \texttt{TargetClass} as `categorical`. This ensures schema compliance and enables CTGAN to embed and conditionally sample from discrete feature spaces.

The model is trained with the following hyperparameters: batch size of 500, learning rate of $2 \times 10^{-4}$, and 300 training epochs. The training objective follows the WGAN-GP formulation to stabilise convergence. During each iteration, CTGAN samples a conditional vector from the discrete columns, embeds categorical features, and concatenates them with Gaussian noise to form the generator input. The discriminator receives both real and synthetic data with their associated conditions, ensuring the generator learns contextually valid feature-label combinations.

    \item \textbf{Gaussian Noise Injection for Data Augmentation}
    
    In this approach, Gaussian noise $\mathcal{N}(\mu, \sigma^2)$ is added to the numerical features during training, following the methodology outlined in \cite{inproceedings}. This noise injection serves to regularise training, preventing the model from memorising training patterns and improving its generalisation ability.
    
    \paragraph{Dynamic Noise Adjustment}
    To ensure effective augmentation, noise parameters are dynamically tuned based on feature distributions. Specifically, for each feature $x_i$, the noise follows:
    \begin{equation}
    x_i' = x_i + \epsilon, \quad \epsilon \sim \mathcal{N}(0, \sigma^2_{x_i})
    \end{equation}
    where $\sigma_{x_i}$ is adaptively set as a fraction of the empirical standard deviation of $x_i$.
\end{itemize}

\section{Evaluation and Validation}
\label{sec:eval}

\subsection{Experimental Protocol}
\label{sec:eval:cv}

The effectiveness of synthetic data augmentation, generated using WGAN, CTGAN, a diffusion model, and Gaussian noise augmentation, is assessed through two approaches: \textbf{semi-synthetic} and \textbf{full-synthetic}. The \textit{semi-synthetic} approach augments the original dataset with synthetic observations, while the \textit{full-synthetic} approach replaces real data entirely with synthetic samples. The impact of these approaches is evaluated based on the following criteria:

\begin{itemize}
    \item \textbf{Distribution Fidelity:} The similarity between synthetic and real data distributions is measured using Kullback-Leibler (KL)~\cite{kullback1951} and Jensen-Shannon (JS)~\cite{menendez1997jensen} to assess how well synthetic data preserves the statistical properties of the original dataset.
    
    \item \textbf{Utility:} Classification models trained on semi-synthetic and full-synthetic datasets are compared against those trained on real data alone. Performance metrics are analysed to determine whether data augmentation enhances predictive generalisation.
    
    \item \textbf{Privacy Robustness:} The degree to which synthetic data protects privacy is assessed using adversarial attack simulations, including \ac{MIA} and Reconstruction Attacks. 
\end{itemize}


\subsubsection{Fidelity}
\label{sec:fidelity}

\begin{figure}[h]
    \centering
    \includegraphics[width=0.45\textwidth]{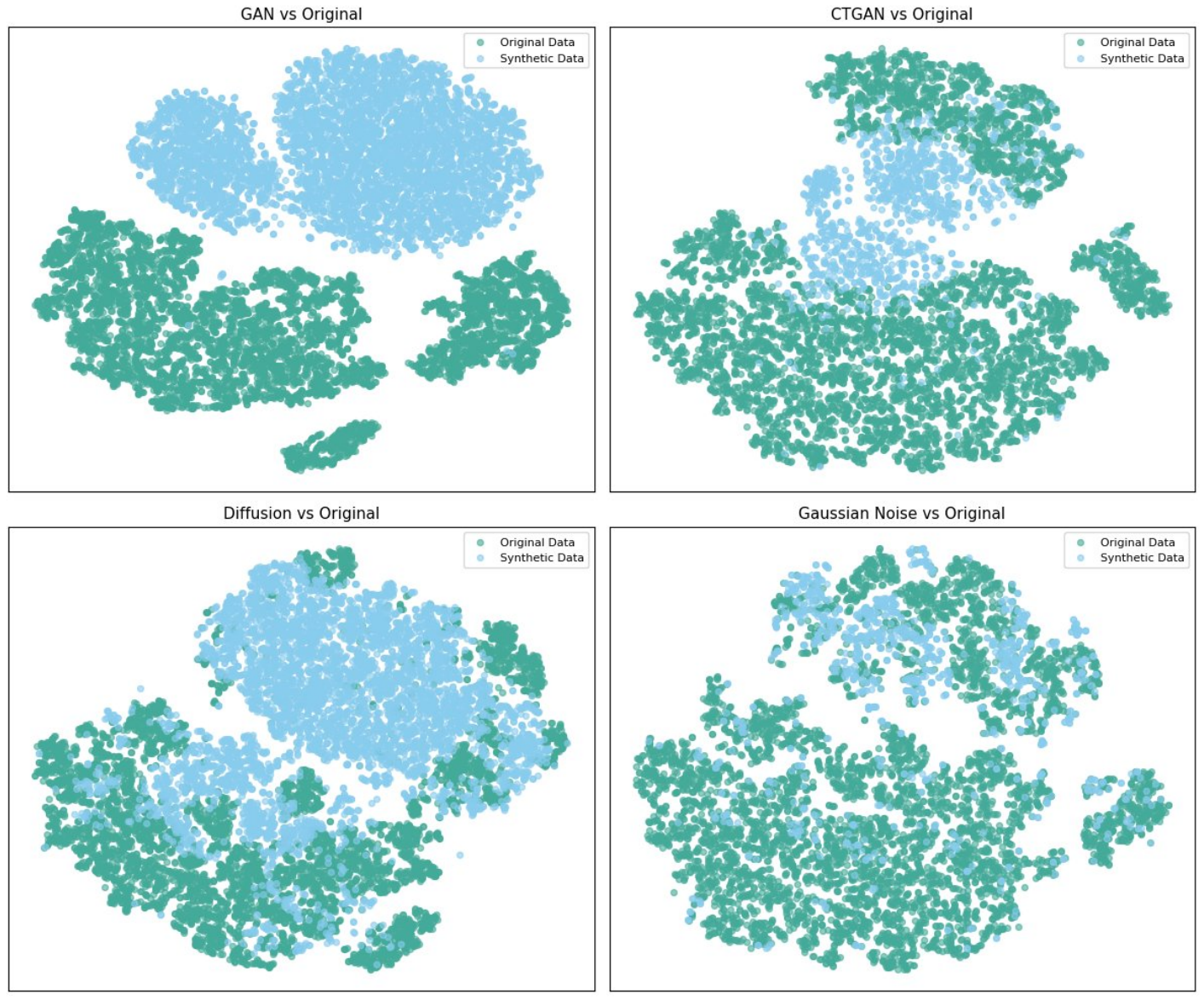}
    \caption{t\mbox{-}SNE (perplexity 30) of real vs.~synthetic samples.
    Colours denote generator + synthesis regime.}
    \label{fig:tsne}
\end{figure}

\paragraph{Visual inspection.}
Figure~\ref{fig:tsne} overlays the t\mbox{-}SNE embeddings of the original records and eight synthetic variants.  Large overlaps (e.g., Gaussian noise augmentation vs Original) indicate that synthetic points populate the same manifold regions as the real data, whereas visible cluster gaps reveal distributional drift (notably for the WGAN synthetic samples).

\begin{table}[h]
\centering
\caption{Fidelity and utility metrics across synthetic data types. Best KL/JS in bold, worst underlined.}
\label{tab:combined_divergence_utility}
\scriptsize
\begin{tabularx}{\columnwidth}{Xcccc}
\toprule
\textbf{Synthetic data} & \textbf{KL} & \textbf{JS} & \textbf{Best Classifier} & \textbf{Macro‑F1 (\%)} \\
\midrule
Real         & ---- & ---- & XGB & 67.5 ± 3.8  \\
\midrule
WGAN Semi‑synthetic          & 2.68 & 0.083 & XGB & 63.4 ± 5.3 \\
WGAN Full‑synthetic          & 6.12 & 0.221 & KNN & 50.7 ± 4.2 \\
\midrule
\textbf{CTGAN Semi‑synth.} & \textbf{0.16} & \textbf{0.014} & XGB & 68.3 ± 5.4 \\
CTGAN Full‑synth.        & 0.44 & 0.045 & RF & 82.5 ± 4.3 \\
\midrule
Diffusion Semi‑synth.    & 0.74 & 0.021 & RF & 73.0 ± 5.1 \\
Diffusion Full‑synth.    & 1.44 & 0.054 & SVM & 88.2 ± 3.0 \\
\midrule
Noise Baseline Semi         & 0.27 & 0.007 & KNN & 75.7 ± 5.1 \\
Noise Baseline Full         & 0.52 & 0.016 & KNN & 81.6 ± 2.9 \\
\bottomrule
\end{tabularx}
\end{table}

\paragraph{Divergence metrics.}
To move beyond visual impression, we compute the Kullback–Leibler (KL) and Jensen–Shannon (JS) divergences between the multivariate Gaussian kernel‑density estimates of the real data and each synthetic distribution (Table~\ref{tab:combined_divergence_utility}).\footnote{Following Xu \emph{et al.} \cite{xu2019ctgan}, we estimate the probability density of each dataset with a Gaussian KDE whose bandwidth is chosen by Scott’s rule \cite{scott1992density}, i.e.\ $h_j = \sigma_j n^{-1/(d+4)}$.
For our real data ($n=1117$, $d=24$) this yields $h_j \approx 0.78\,\sigma_j$ for every feature.}

\begin{table*}[h]
\centering
\caption{Macro‑F1 (\%) on each dataset.
         \(\uparrow\) / \(\downarrow\): significant improvement / drop
         vs.\ real data ($p<0.05$).\hfill}
\label{tab:utility_real_vs_synth}
\small
\setlength{\tabcolsep}{5pt}
\begin{tabular}{@{}lccccc@{}}
\toprule
\textbf{Dataset} & \textbf{RF} & \textbf{KNN} & \textbf{SVM} & \textbf{DT} & \textbf{XGB} \\
\midrule
Real (baseline)              & 66.5 $\pm$ 4.1 & 62.1 $\pm$ 3.2 & 65.0 $\pm$ 4.0 & 62.0 $\pm$ 3.0 & \textbf{67.5} $\pm$ 3.8 \\
\addlinespace[2pt]
WGAN Semi‑Synth.              & 61.0 $\pm$ 5.3\,\(\downarrow\) & 55.3 $\pm$ 5. \,\(\downarrow\) & 61.6 $\pm$ 4.1 & 60.0 $\pm$ 5.5 & \textbf{63.4} $\pm$ 5.3 \\
WGAN Full‑Synth.              & 42.6 $\pm$ 2.6\,\(\downarrow\) & \textbf{50.7} $\pm$ 4.2\,\(\downarrow\) & 50.5 $\pm$ 3.5\,\(\downarrow\) & 48.6 $\pm$ 3.4\,\(\downarrow\) & 45.7 $\pm$ 2.9\,\(\downarrow\) \\
CTGAN Semi‑Synth.            & 67.6 $\pm$ 4.2 & 61.1 $\pm$ 4.9 & 67.5 $\pm$ 4.4 & 62.3 $\pm$ 4.8 & \textbf{68.3} $\pm$ 5.4 \\
CTGAN Full‑Synth.            & \textbf{82.5} $\pm$ 4.3\,\(\uparrow\) & 75.1 $\pm$ 3.4\,\(\uparrow\) & 80.1 $\pm$ 4.2\,\(\uparrow\) & 70.2 $\pm$ 3.6\,\(\uparrow\) & 80.0 $\pm$ 3.1\,\(\uparrow\) \\
Diffusion Semi‑Synth.        & 73.0 $\pm$ 5.1 & 62.5 $\pm$ 3.4 & 71.5 $\pm$ 3.6\,\(\downarrow\) & 69.9 $\pm$ 3.8\,\(\downarrow\) & 72.0 $\pm$ 4.  \\
Diffusion Full‑Synth.        & 82.9 $\pm$ 3.1\,\(\uparrow\) & 68.8 $\pm$ 3.0\,\(\uparrow\) & \textbf{88.2} $\pm$ 3.0\,\(\uparrow\) & 79.2 $\pm$ 4.0\,\(\uparrow\) & 87.4 $\pm$ 4.0\,\(\uparrow\) \\
Noise Semi‑Synth.   & 74.8 $\pm$ 5.9\,\(\uparrow\) & 75.7 $\pm$ 5.1\,\(\uparrow\) & 73.6 $\pm$ 4.6\,\(\uparrow\) & 69.3 $\pm$ 2.7\,\(\uparrow\) & 75.5 $\pm$ 5.8\,\(\uparrow\) \\
Noise Full‑Synth.            & 76.0 $\pm$ 3.7\,\(\uparrow\) & \textbf{81.6} $\pm$ 2.9\,\(\uparrow\) & 77.2 $\pm$ 4.4\,\(\uparrow\) & 72.6 $\pm$ 3.0\,\(\uparrow\) & 79.2 $\pm$ 3.0\,\(\uparrow\) \\

\bottomrule
\end{tabular}
\end{table*}

Both metrics reward exact overlap (0.0 is best), but JS is symmetric and bounded, making it easier to compare across methods.


\noindent
\textbf{Observations.}
CTGAN Semi‑synthetic achieves the lowest divergence on both metrics, indicating that a modest proportion of real samples is sufficient for that generator to learn high‑order structure.  Diffusion models come second, followed closely by the calibrated Gaussian‑noise injection.
WGAN Full‑synthetic is an outlier with a KL of 6.1, evidence of model collapse and excessive variance inflation.

\paragraph{Summary‑statistics parity.}
Beyond global divergence metrics, we assess fidelity by comparing means, skewness, and kurtosis of 24 numerical features. \textbf{CTGAN Semi‑Synthetic} best replicates original distributions, preserving asymmetry and tail behavior in complex features like \texttt{Peak\_to\_Mean\_Ratio}. In contrast, \textbf{WGAN Full‑Synthetic} flattens key statistics. Diffusion and noise-based models preserve central moments but tend to exaggerate tails. These findings highlight the importance of matching higher-order moments for faithful data synthesis.

\paragraph{Overall fidelity score.}
Combining (i) low KL/JS, (ii) moment parity, and (iii) the qualitative t\mbox{-}SNE overlap, we conclude that \textbf{CTGAN Semi‑synthetic delivers the highest distributional fidelity}, closely followed by Diffusion Semi‑synthetic. WGAN full‑synthetic approach, while visually plausible, deviates substantially in tail behaviour and is therefore \emph{not recommended} for downstream tasks that rely on accurate peak or burst modelling.

\subsection{Utility}

\paragraph{Performance Comparison}
 Table~\ref{tab:utility_real_vs_synth} summarises the classification performance across the different augmentation strategies, highlighting that diffusion-based full-synthetic data yields the highest Macro-F1 score.

\paragraph{Significance Testing}
\label{sec:eval:sig}
To compare synthetic against real data, we apply two paired tests across the outer‑fold macro‑F1 vectors: (i) two‑sided Wilcoxon signed‑rank \cite{wilcoxon1945}, and (ii) paired Student \(t\)-test \cite{student1908}. The null hypothesis states that the mean performance difference is zero. $p<0.05$ (after Holm–Bonferroni correction across five classifiers) flags a \emph{significant utility drop}.  Tables~\ref{tab:combined_divergence_utility} and~\ref{tab:utility_real_vs_synth} showcase the results.

To obtain an unbiased estimate of downstream utility, we adopt the fixed \emph{nested} cross-validation pipeline summarised in Algorithm ~\ref{alg:nestedcv}:

\begin{algorithm}[h]
\caption{Nested $5 \times 3$ cross-validation (macro-$\mathrm{F1}$).}
\label{alg:nestedcv}
\begin{algorithmic}[1]
\REQUIRE data $(X, y)$, classifiers $\mathcal{C}$, outer $K=5$, inner $k=3$
\FOR{each classifier $h \in \mathcal{C}$}
    \FOR{outer fold $i = 1$ to $K$}
        \STATE Split $(X, y)$ into $\text{train}_i$, $\text{test}_i$
        \STATE \textbf{Hyper-parameter search}: RandomisedSearchCV on $\text{train}_i$ using inner $k$ folds
        \STATE Fit best $h^*$ on $\text{train}_i$
        \STATE Evaluate on $\text{test}_i$; store macro-$\mathrm{F1}_i$
    \ENDFOR
    \STATE Report $\mu = \operatorname{mean}(\mathrm{F1}_1 \dots \mathrm{F1}_K)$,\;
           $\sigma$, and $95\%$ CI $= 1.96 \sigma / \sqrt{K}$
\ENDFOR
\end{algorithmic}
\end{algorithm}

\paragraph{Hyper‑parameter search.}
We draw \(n_{\!\text{iter}}=10\) random configurations per classifier 5 are evaluated in each outer CV fold) from the grids listed in Table~\ref{tab:grids}. 

\begin{table}[h]
\centering
\caption{Randomised search spaces for the baseline classifiers.
         “\(a{:}b{:}c\)” follows \texttt{range(a,\,c{+}1,\,b)}.}
\label{tab:grids}
\small
\begin{tabularx}{\linewidth}{@{}lX@{}}
\toprule
\textbf{Model} & \textbf{Parameter grid} \\ \midrule
RF  & \texttt{n\_estimators}\,$\in$\{100,\,300,\,500\};\,
       \texttt{max\_depth}\,$\in$\{None,\,10,\,20\};\,
       \texttt{min\_samples\_split}\,$\in$\{2,\,5\} \\[2pt]
KNN & \texttt{n\_neighbors}\,=\,3{:}2{:}15;\,
       \texttt{weights}\,$\in$\{uniform,\,distance\} \\[2pt]
SVM & \texttt{C}\,$\in$\{1,\,10\};\,
       \texttt{kernel}\,=\,\texttt{rbf};\,
       \texttt{gamma}\,=\,\texttt{scale} \\[2pt]
XGB & \texttt{n\_estimators}\,$\in$\{100,\,300,\,500\};\,
       \texttt{max\_depth}\,$\in$\{3,\,6,\,10\};\,
       \texttt{learning\_rate}\,$\in$\{0.01,\,0.1,\,0.3\};\,
       \texttt{subsample}\,$\in$\{0.8,\,1.0\} \\[2pt]
DT  & \texttt{max\_depth}\,$\in$\{None,\,10,\,20,\,30\};\,
       \texttt{criterion}\,$\in$\{gini,\,entropy\} \\ \bottomrule
\end{tabularx}
\end{table}

\subsection{Privacy Robustness}
\label{sec:privacy}

Protecting individual records from disclosure is a prerequisite when synthetic data are released or used to train downstream models.  We study the canonical \emph{membership inference attack} (MIA) under the strong \textbf{posterior‑only black‑box} threat model: The adversary observes the prediction vector of the target model for an arbitrary query but has no access to gradients, weights, or training loss.

\subsubsection{MIA}

\begin{itemize}
  \item \textbf{Shadow ensemble.}  For each dataset we fit $n_\text{shadow}=5$ Shadow models with heterogeneous architectures (Random Forest and 2‑layer MLP).  Hyper‑parameters and seeds follow the public YAML file in our code release.
  \item \textbf{Attack features.}  From every shadow inference we extract the maximum posterior probability (“\texttt{max‑prob}”), a standard and practically obtainable signal~\cite{shokri2017membershipinferenceattacksmachine}.
  \item \textbf{Attack classifiers.}  We train both a Random Forest (200 trees) and an MLP (32–32 units) on the aggregated shadow‑generated attack dataset, using a \mbox{10\%} hold‑out portion for early stopping. Only the \emph{stronger} attacker per setting is reported.
  \item \textbf{Repeats and confidence intervals.} The entire pipeline is repeated for five independent seeds.  We report the mean attack AUC together with the \emph{two‑sided 95\% t‑interval}.

\end{itemize}

\begin{table}[h]
\centering
\caption{Membership inference attack results (higher AUC = more vulnerable). Parentheses show 95\% confidence interval.}
\label{tab:mia_auc}
\scriptsize
\begin{tabularx}{\columnwidth}{Xc}
\toprule
\textbf{Synthetic data} & \textbf{Shadow MIA AUC} \\
\midrule
WGAN Semi‑synthetic          & 0.62\;(0.60, 0.63) \\
WGAN Full‑synthetic          & 0.62\;(0.61, 0.64) \\
CTGAN Semi‑synthetic        & 0.63\;(0.61, 0.64) \\
CTGAN Full‑synthetic        & 0.64\;(0.62, 0.65) \\
Diffusion Semi‑synthetic    & 0.62\;(0.61, 0.63) \\
Diffusion Full‑synthetic    & 0.64\;(0.62, 0.65) \\
Noise Semi-synthetic        & 0.63\;(0.61, 0.64) \\
Noise Full-synthetic         & 0.61\;(0.60, 0.63) \\
\midrule
\textbf{Original data}      & \textbf{0.65\;(0.64, 0.66)} \\
\bottomrule
\end{tabularx}
\end{table}

\paragraph{Discussion.}
Across all eight synthetic configurations illustrated in Table~\ref{tab:mia_auc}, the attack AUC stays in the narrow range \mbox{0.61–0.64}, only slightly above random guessing.  Differences between \emph{semi‑} and \emph{full‑synthetic} variants of the same generator never exceed $\Delta$AUC $=0.03$.  The Gaussian‑noise approach offers the strongest privacy (\mbox{0.61})).  In contrast, diffusion‑based full synthesis yields the best trade‑off: a modest AUC of 0.64 while increasing the target’s predictive accuracy.

These findings suggest that the synthesis pipeline, in its current form, does not adequately safeguard against privacy leakage, prompting a more detailed examination of its susceptibility to feature reconstruction attacks thereafter

\subsubsection{Reconstruction attack}
\label{sec:reconstruction}

\paragraph{Threat model.}
An adversary receives the released \emph{synthetic table}. Using only the published attributes, it trains a regression model that tries to \emph{reconstruct} a hidden target feature of interest, in our case, its \emph{average consumption}.
Successful reconstruction reveals fine‑grained private behaviour.

\paragraph{Attack procedure.}

\begin{enumerate}[leftmargin=*,nosep,label=(\arabic*)]
  \item \textbf{Model sweep.} Train five regressors \{Random Forest, Gradient‑Boosting, Ridge, Lasso, MLP\} on the synthetic data. Pick the model with the lowest mean‑squared error (MSE) on a held‑out \emph{real} test set.  This “best of sweep’’ yields an upper bound on what a resourceful adversary could achieve.
  \item \textbf{Real‑data upper bound.} Train the same Random Forest directly on the real table; its error is the maximum leak possible if the raw data were published.
  \item \textbf{Privacy metrics.} \emph{Privacy gap} $\Delta = \text{MSE}_{\text{noise}} - \text{MSE}_{\text{syn}}$ and \emph{privacy‑risk score} $\text{PRS}=\Delta / (\text{MSE}_{\text{noise}}-\text{MSE}_{\text{real}})$ (ratio in $[0,1]$; higher $\Rightarrow$ bigger leak).
\end{enumerate}

\begin{table}[h]
\centering
\caption{Reconstruction results (lower MSE / PRS = safer). Best value per column in \textbf{bold}, worst underlined.}
\label{tab:reconstruction_attack}
\scriptsize
\begin{tabularx}{\columnwidth}{Xccccc}
\toprule
\textbf{Synthetic data} &
\textbf{Best model} &
\textbf{MSE\,($\times10^{-3}$)} &
\textbf{$\rho$} &
\textbf{$R^{2}$} &
\textbf{PRS} \\
\midrule
WGAN Semi‑synthetic            & RF    & 0.18 & 0.996 & 0.990 & 0.99 \\
WGAN Full‑synthetic & GB    & \ 8.30 & 0.760 & 0.561 & 0.62 \\
CTGAN Semi‑synthetic         & RF    & 0.27 & 0.994 & 0.986 & 0.99 \\
\textbf{CTGAN Full‑synthetic}& \textbf{Lasso} & \textbf{19.1} & \textbf{- 0.022} & \textbf{--0.010} & \textbf{0.16} \\
Diffusion Semi‑synthetic     & GB    & 0.07 & 0.998 & 0.996 & 0.99 \\
Diffusion Full‑synthetic      & Ridge & 0.35 & 0.994 & 0.982 & 0.98 \\
Noise Semi-Synthetic         & GB    & 0.00 & 1.000 & 0 999 & 1. 0 \\
Noise Full-synthetic        & GB    & 0.05 & 0.999 & 0 997 & 1. 0 \\
\bottomrule
\end{tabularx}
\end{table}

From table ~\ref{tab:reconstruction_attack} we extract the following findings:
\begin{enumerate}[leftmargin=*]
  \item \textbf{Semi‑synthetic sets leak almost as much as the real data.} Their errors are three orders of magnitude smaller than the noise baseline (MSE $\le0.27\!\times\!10^{-3}$) and correlations exceed $0.99$, giving PRS \(\approx1\).
  \item \textbf{CTGAN Full‑synthetic offers the strongest protection.} Error jumps two orders of magnitude (MSE $=19.1\!\times\!10^{-3}$), correlation vanishes, and PRS drops to 0.16—\emph{84 \% less leak} than raw data. Diffusion Full‑synthetic keeps good fidelity (MSE $=0.35$) but still leaks almost as much as noise‑shuffled data (PRS 0.98).
  \item \textbf{WGAN Full‑synthetic} partially mitigates risk (PRS 0.62) but still leaves a strong linear correlation between synthetic and real data (\(\rho=0.76\)).
\end{enumerate}

\paragraph{Observations. }
When the secret attribute is a fine‑grained load profile,
\textbf{CTGAN full‑synthetic is the only generator that meaningfully degrades reconstruction attacks} without resorting to external noise addition. Combining these results with the MIA study, we recommend CTGAN full synthesis for any public release that prioritises privacy; diffusion full synthesis remains acceptable for internal analytics where some residual leakage is tolerable.

\subsubsection{Privacy–Utility Pareto}

\begin{figure}[h]
    \centering
    \includegraphics[width=0.5\textwidth]{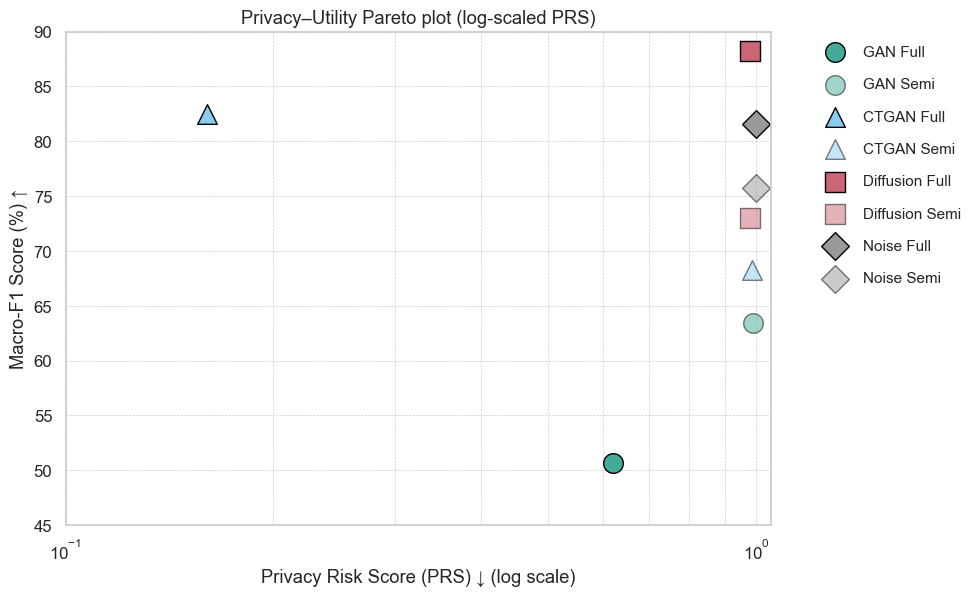}
    \caption{Privacy–Utility Pareto plot. Marker shape encodes model type (GAN: circle, CTGAN: triangle, Diffusion: square, Noise: diamond). Color encodes data type (Full: darker; Semi: lighter). 
}
     \label{fig:pareto}
\end{figure}

Figure~\ref{fig:pareto} presents the Privacy–Utility Pareto frontier, mapping each generative strategy along two axes: privacy, quantified by the Privacy Risk Score (PRS, x-axis, lower is better), and utility, measured by macro-F1 classification performance (y-axis, higher is better). 

The figure highlights clear trade-offs, where structured generative models, especially CTGAN and Diffusion, define the Pareto frontier, confirming CTGAN full synthesis offers the strongest defence, while diffusion provides the best compromise between privacy and predictive utility.

\vspace{-1cm}

\section{Discussion}
\subsection{Model Effectiveness}
Classifier performance varies considerably across synthetic data strategies. While \textbf{SVM achieves the highest macro‑F1 score overall (88.2\%) on Diffusion Full‑Synthetic data}, it is not universally the best across all datasets. For instance, \textbf{Random Forest} is best on CTGAN Full‑Synthetic (82.5\%), \textbf{XGBoost} on Diffusion Full‑Synthetic (88.2\%), and \textbf{KNN} performs best for both Noise Baseline Full (81.6\%) and \ac{WGAN} Full‑Synthetic (50.7\%). This suggests that no single model is optimal across all synthetic settings, and model selection should be data-specific. Notably, Decision Trees consistently underperform, showing the lowest scores for both real and synthetic data, likely due to their higher variance and susceptibility to overfitting.

\subsection{Impact of Synthetic Data on Performance}

The utility of classifiers trained on synthetic data reveals clear distinctions between generation methods. \textbf{Diffusion Full‑Synthetic data consistently achieves the highest utility across classifiers}, with macro‑F1 scores reaching up to 88.2\% (SVM), a substantial improvement over the real data baseline (67.5\%). CTGAN Full‑Synthetic also demonstrates strong performance (82.5\%, RF), outperforming its Semi‑Synthetic counterpart. This supports findings by \cite{synthetic_residential_load}, which demonstrated that WGAN-based synthetic data enhances forecasting performance when integrated with empirical data. However, our work differs in that we systematically compare multiple generative approaches for tariff responsiveness classification rather than load forecasting.
Additionally, while authors in \cite{sensor_data_augmentation} showed that Gaussian noise augmentation improves forecasting accuracy, our results indicate that diffusion models provide a greater performance boost in predictive modelling for tariff recommendation. This suggests that more structured synthetic data generation methods can be preferable to noise-based augmentation in classification tasks.

\subsection{Privacy Considerations}
Our study confirms that full-synthetic datasets provide stronger defences against Membership Inference and Reconstruction Attacks compared to real data. This aligns with prior research on synthetic data privacy, such as \cite{chai2024faradaysyntheticsmartmeter}, which demonstrated that VAE-generated load profiles reduce privacy risks while preserving data utility.

However, unlike authors in \cite{ctgan_short_term_forecasting}, which focused on hybrid models combining real and synthetic data for forecasting, our evaluation explicitly quantifies privacy risks using privacy attacks. 

\subsection{Answering Research Questions}
\paragraph{\textbf{RQ1} - Model Effectiveness} Our results confirm that SVM and XGBoost achieve the highest classification performance, demonstrating strong predictive capability with a macro-F1 score of 0.88. This aligns with prior work on classification using smart grid data \cite{petrlik2022electricity, mhaske2022efficient} but extends it to tariff responsiveness classification.

\paragraph{\textbf{RQ2} - Impact of Synthetic Data} Diffusion Full‑Synthetic data provides the most significant utility gains, outperforming real data by over 20 macro‑F1 points in some cases. Unlike \cite{sensor_data_augmentation}, which used Gaussian noise for load forecasting, our study shows that structured generative models yield superior results for classification.


\paragraph{\textbf{RQ3} - Privacy Protection}
Across all generators, membership inference attacks achieve AUCs no higher than 0.64, replicating the near‑random leakage observed for diffusion backbones by Duan \& Liang \cite{wu2025winning}. For reconstruction, however, the risk diverges: our \emph{CTGAN full‑synthetic} attains the lowest privacy–risk score (PRS = 0.16), confirming the superior robustness reported by Alshantti et al.\ \cite{alshantti2024privacyreidentificationattackstabular}, whereas WGAN‑GP variants suffer the highest leak, consistent with Hyeong et al.’s empirical ranking \cite{hyeong2022empiricalstudymembershipinference}.

\section{Conclusion}
This study evaluated the impact of different synthetic data generation architectures on the trade-off between utility and privacy in the context of household classification for \ac{dToU} tariffs. We benchmarked four generation strategies, \ac{WGAN}, CTGAN, Diffusion Models, and Gaussian noise augmentation, under both full-synthetic and semi-synthetic regimes. Utility was measured via macro‑F1 score, fidelity via KL/JS divergence, and privacy via \ac{MIA} and feature reconstruction.

Our findings, as illustrated in Figure \ref{fig:pareto}, provide empirical support for the hypothesis that \textbf{the structural design of synthetic data generators significantly affects both privacy and utility}. Specifically:
\begin{itemize}
    \item \textbf{Diffusion Full‑Synthetic data}, characterized by step-wise denoising and timestep conditioning, yielded the \emph{highest utility}.
    \item \textbf{CTGAN Full‑Synthetic}, which employs conditional generation and tabular-specific modeling, achieved \emph{the best privacy protection} (lowest PRS = 0.16), albeit with slightly lower utility than diffusion.
    \item In contrast, \textbf{WGAN Full‑Synthetic} exhibited \emph{the weakest performance} in both utility and privacy, suggesting that architectural limitations such as mode collapse and lack of conditioning impair generalisation and increase leakage risks.
    \item \textbf{Gaussian noise}, while showing high utility, closely mimicked real data (low KL/JS), resulting in higher reconstruction scores and reduced privacy guarantees.
\end{itemize}

These results confirm that more structured architectures (e.g., Diffusion and CTGAN) are better suited for privacy-preserving synthetic data generation in smart grid applications.

\section{Threats to Validity}

\textbf{Label construction bias.} The ground-truth labels for responsiveness are derived from unsupervised dimensionality reduction and quantile-based thresholding. While this procedure is grounded in behavioural features, it may not align perfectly with economic responsiveness or actual behavioural change. The absence of external validation data, such as expert annotations or outcomes from real-world dToU interventions, limits our ability to confirm the semantic validity of these labels.

\textbf{Contextual scope limitations.} Our analysis is grounded in a single empirical dataset from the UK Power Networks trial. While this dataset is representative of real smart meter deployments and tariff experiments, its demographic, regional, and temporal scope may not generalize to other electricity markets or consumption behaviours. Caution is warranted when extrapolating our results to other jurisdictions or pricing schemes.

\textbf{Limitations of empirical privacy evaluation.} Our privacy assessment relies on empirical adversaries, membership inference and reconstruction attacks, without incorporating formal guarantees. These attacks are practical and widely used but may underestimate leakage against more adaptive or theoretically grounded threats.

\section{Future Work}
While this research establishes the feasibility of using synthetic data for predicting household responsiveness to \ac{dToU} tariffs, several avenues exist for advancing the generative modelling framework. One key direction is the integration of time-aware generative architectures that better capture temporal dependencies and consumption dynamics. For instance, TimeGAN~\cite{timegan} combines autoregressive recurrent models with adversarial training to preserve both temporal ordering and feature relationships, offering a strong foundation for sequence-level generation in smart meter data. More recent diffusion-based time-series models such as Time-series Dense Encoder (TiDE)~\cite{tidemodel} and Conditional Score-based Diffusion Models for Probabilistic Time Series Imputation (CSDI)~\cite{tashiro2021csdi} introduce transformer-based architectures and conditional score-based imputation, which may yield further improvements in fidelity and controllability of temporal patterns. In parallel, incorporating explicit causal structure into the generative process could enable more principled simulation of consumption responses under varying tariff schemes. 

A current limitation of our work lies in the use of unsupervised methods to construct binary responsiveness labels. While PCA-based thresholds offer a principled starting point, they may not fully reflect real-world responsiveness behaviours. Future work should perform a sensitivity analysis across different thresholding strategies and consider triangulating results with expert annotations, user studies, or empirical evidence from tariff response trials \cite{faruqui2010household, privacy_concerns}.

Finally, while our empirical privacy evaluation using membership inference and reconstruction attacks provides practical insights, our current approach does not offer formal guarantees. Future work should explore integrating certified mechanisms, such as Differentially Private Stochastic Gradient Descent (DP-SGD)~\cite{10.1145/2976749.2978318} or the Private Aggregation of Teacher Ensembles (PATE) framework~\cite{papernot2017semisupervisedknowledgetransferdeep}, to establish provable bounds on privacy leakage during synthetic data generation.

By addressing these challenges, future research can further improve the effectiveness and reliability of synthetic data-driven tariff allocation systems.


\clearpage
\bibliography{egbib}

\begin{thebibliography}{38}
\providecommand{\natexlab}[1]{#1}
\providecommand{\url}[1]{\texttt{#1}}
\expandafter\ifx\csname urlstyle\endcsname\relax
  \providecommand{\doi}[1]{doi: #1}\else
  \providecommand{\doi}{doi: \begingroup \urlstyle{rm}\Url}\fi

\bibitem[Abadi et~al.(2016)Abadi, Chu, Goodfellow, McMahan, Mironov, Talwar, and Zhang]{10.1145/2976749.2978318}
M.~Abadi, A.~Chu, I.~Goodfellow, H.~B. McMahan, I.~Mironov, K.~Talwar, and L.~Zhang.
\newblock Deep learning with differential privacy.
\newblock In \emph{Proceedings of the 2016 ACM SIGSAC Conference on Computer and Communications Security}, CCS '16, page 308–318, New York, NY, USA, 2016. Association for Computing Machinery.
\newblock ISBN 9781450341394.
\newblock \doi{10.1145/2976749.2978318}.
\newblock URL \url{https://doi.org/10.1145/2976749.2978318}.

\bibitem[Albert and Rajagopal(2013)]{6545387}
A.~Albert and R.~Rajagopal.
\newblock Smart meter driven segmentation: What your consumption says about you.
\newblock \emph{IEEE Transactions on Power Systems}, 28\penalty0 (4):\penalty0 4019--4030, 2013.
\newblock \doi{10.1109/TPWRS.2013.2266122}.

\bibitem[Alshantti et~al.(2024)Alshantti, Rasheed, and Westad]{alshantti2024privacyreidentificationattackstabular}
A.~Alshantti, A.~Rasheed, and F.~Westad.
\newblock Privacy re-identification attacks on tabular gans, 2024.
\newblock URL \url{https://arxiv.org/abs/2404.00696}.

\bibitem[Beckel et~al.(2014)Beckel, Sadamori, Staake, and Santini]{BECKEL2014397}
C.~Beckel, L.~Sadamori, T.~Staake, and S.~Santini.
\newblock Revealing household characteristics from smart meter data.
\newblock \emph{Energy}, 78:\penalty0 397--410, 2014.
\newblock ISSN 0360-5442.
\newblock \doi{https://doi.org/10.1016/j.energy.2014.10.025}.
\newblock URL \url{https://www.sciencedirect.com/science/article/pii/S0360544214011748}.

\bibitem[Chai and Chadney(2024)]{chai2024faradaysyntheticsmartmeter}
S.~Chai and G.~Chadney.
\newblock Faraday: Synthetic smart meter generator for the smart grid, 2024.
\newblock URL \url{https://arxiv.org/abs/2404.04314}.

\bibitem[Faruqui and Sergici(2010)]{faruqui2010household}
A.~Faruqui and S.~Sergici.
\newblock Household response to dynamic pricing of electricity: a survey of 15 experiments.
\newblock \emph{Journal of Regulatory Economics}, 38\penalty0 (2):\penalty0 193--225, 2010.

\bibitem[Freier and {von Loessl}(2022)]{dynamic_tariffs_design}
J.~Freier and V.~{von Loessl}.
\newblock Dynamic electricity tariffs: Designing reasonable pricing schemes for private households.
\newblock \emph{Energy Economics}, 112:\penalty0 106146, 2022.
\newblock ISSN 0140-9883.
\newblock \doi{https://doi.org/10.1016/j.eneco.2022.106146}.
\newblock URL \url{https://www.sciencedirect.com/science/article/pii/S0140988322003012}.

\bibitem[Goodfellow et~al.(2020)Goodfellow, Pouget-Abadie, Mirza, Xu, Warde-Farley, Ozair, Courville, and Bengio]{goodfellow2020generative}
I.~Goodfellow, J.~Pouget-Abadie, M.~Mirza, B.~Xu, D.~Warde-Farley, S.~Ozair, A.~Courville, and Y.~Bengio.
\newblock Generative adversarial networks.
\newblock \emph{Communications of the ACM}, 63\penalty0 (11):\penalty0 139--144, 2020.

\bibitem[Gulrajani et~al.(2017)Gulrajani, Ahmed, Arjovsky, Dumoulin, and Courville]{gulrajani2017improvedtrainingwassersteingans}
I.~Gulrajani, F.~Ahmed, M.~Arjovsky, V.~Dumoulin, and A.~Courville.
\newblock Improved training of wasserstein gans, 2017.
\newblock URL \url{https://arxiv.org/abs/1704.00028}.

\bibitem[Guo and Weeks(2022)]{dynamic_tariffs_response}
B.~Guo and M.~Weeks.
\newblock Dynamic tariffs, demand response, and regulation in retail electricity markets.
\newblock \emph{Energy Economics}, 106:\penalty0 105774, 2022.
\newblock ISSN 0140-9883.
\newblock \doi{https://doi.org/10.1016/j.eneco.2021.105774}.
\newblock URL \url{https://www.sciencedirect.com/science/article/pii/S0140988321006149}.

\bibitem[Gupta(2023)]{tidemodel}
R.~e.~a. Gupta.
\newblock Tide: Time-series diffusion for forecasting and imputation.
\newblock \emph{arXiv preprint arXiv:2305.13395}, 2023.

\bibitem[Ho et~al.(2020)Ho, Jain, and Abbeel]{ho2020denoisingdiffusionprobabilisticmodels}
J.~Ho, A.~Jain, and P.~Abbeel.
\newblock Denoising diffusion probabilistic models, 2020.
\newblock URL \url{https://arxiv.org/abs/2006.11239}.

\bibitem[Hyeong et~al.(2022)Hyeong, Kim, Park, and Jajodia]{hyeong2022empiricalstudymembershipinference}
J.~Hyeong, J.~Kim, N.~Park, and S.~Jajodia.
\newblock An empirical study on the membership inference attack against tabular data synthesis models, 2022.
\newblock URL \url{https://arxiv.org/abs/2208.08114}.

\bibitem[Khorramshahi et~al.(2020)Khorramshahi, Souri, Chellappa, and Feizi]{khorramshahi2020gansvariationalentropyregularizers}
P.~Khorramshahi, H.~Souri, R.~Chellappa, and S.~Feizi.
\newblock Gans with variational entropy regularizers: Applications in mitigating the mode-collapse issue, 2020.
\newblock URL \url{https://arxiv.org/abs/2009.11921}.

\bibitem[Kingma and Ba(2017)]{kingma2017adammethodstochasticoptimization}
D.~P. Kingma and J.~Ba.
\newblock Adam: A method for stochastic optimization, 2017.
\newblock URL \url{https://arxiv.org/abs/1412.6980}.

\bibitem[Kullback and Leibler(1951)]{kullback1951}
S.~Kullback and R.~A. Leibler.
\newblock On information and sufficiency.
\newblock \emph{Annals of Mathematical Statistics}, 22\penalty0 (1):\penalty0 79--86, 1951.
\newblock URL \url{https://www.jstor.org/stable/2236703}.

\bibitem[Liang et~al.(2024)Liang, Wang, and Wang]{synthetic_residential_load}
X.~Liang, Z.~Wang, and H.~Wang.
\newblock Synthetic data generation for residential load patterns via recurrent gan and ensemble method.
\newblock \emph{IEEE Transactions on Instrumentation and Measurement}, 73:\penalty0 1--12, 2024.
\newblock \doi{10.1109/TIM.2024.3480225}.

\bibitem[Maalej and Rebai(2021{\natexlab{a}})]{inproceedings}
A.~Maalej and C.~Rebai.
\newblock Sensor data augmentation strategy for load forecasting in smart grid context.
\newblock pages 979--983, 03 2021{\natexlab{a}}.
\newblock \doi{10.1109/SSD52085.2021.9429417}.

\bibitem[Maalej and Rebai(2021{\natexlab{b}})]{sensor_data_augmentation}
A.~Maalej and C.~Rebai.
\newblock Sensor data augmentation strategy for load forecasting in smart grid context.
\newblock In \emph{2021 18th International Multi-Conference on Systems, Signals \& Devices (SSD)}, pages 979--983, 2021{\natexlab{b}}.
\newblock \doi{10.1109/SSD52085.2021.9429417}.

\bibitem[Men{\'e}ndez et~al.(1997)Men{\'e}ndez, Pardo, Pardo, and Pardo]{menendez1997jensen}
M.~L. Men{\'e}ndez, J.~Pardo, L.~Pardo, and M.~Pardo.
\newblock The jensen-shannon divergence.
\newblock \emph{Journal of the Franklin Institute}, 334\penalty0 (2):\penalty0 307--318, 1997.

\bibitem[Mhaske et~al.(2022)Mhaske, Satam, Londhe, and et~al.]{mhaske2022efficient}
D.~Mhaske, R.~Satam, S.~Londhe, and et~al.
\newblock An efficient electricity theft detection using xgboost.
\newblock \emph{International Journal of Engineering Applied Sciences and Technology}, 6\penalty0 (10):\penalty0 282--287, 2022.

\bibitem[Mirza and Osindero(2014)]{mirza2014conditional}
M.~Mirza and S.~Osindero.
\newblock Conditional generative adversarial nets.
\newblock \emph{CoRR}, abs/1411.1784, 2014.
\newblock URL \url{http://arxiv.org/abs/1411.1784}.

\bibitem[Moon et~al.(2020)Moon, Jung, Park, and Hwang]{ctgan_short_term_forecasting}
J.~Moon, S.~Jung, S.~Park, and E.~Hwang.
\newblock Conditional tabular gan-based two-stage data generation scheme for short-term load forecasting.
\newblock \emph{IEEE Access}, 8:\penalty0 205327--205339, 2020.
\newblock \doi{10.1109/ACCESS.2020.3037063}.

\bibitem[Papernot et~al.(2017)Papernot, Abadi, Úlfar Erlingsson, Goodfellow, and Talwar]{papernot2017semisupervisedknowledgetransferdeep}
N.~Papernot, M.~Abadi, Úlfar Erlingsson, I.~Goodfellow, and K.~Talwar.
\newblock Semi-supervised knowledge transfer for deep learning from private training data, 2017.
\newblock URL \url{https://arxiv.org/abs/1610.05755}.

\bibitem[Petrlik et~al.(2022)Petrlik, Lezama, Rodriguez, and et~al.]{petrlik2022electricity}
I.~Petrlik, P.~Lezama, C.~Rodriguez, and et~al.
\newblock Electricity theft detection using machine learning.
\newblock \emph{International Journal of Advanced Computer Science and Applications}, 13\penalty0 (12):\penalty0 420--428, 2022.

\bibitem[Scott(1992)]{scott1992density}
D.~W. Scott.
\newblock \emph{Multivariate Density Estimation: Theory, Practice and Visualization}.
\newblock Wiley, 1992.

\bibitem[Shokri et~al.(2017{\natexlab{a}})Shokri, Stronati, Song, and Shmatikov]{shokri2017membership}
R.~Shokri, M.~Stronati, C.~Song, and V.~Shmatikov.
\newblock Membership inference attacks against machine learning models.
\newblock In \emph{2017 IEEE symposium on security and privacy (SP)}, pages 3--18. IEEE, 2017{\natexlab{a}}.

\bibitem[Shokri et~al.(2017{\natexlab{b}})Shokri, Stronati, Song, and Shmatikov]{shokri2017membershipinferenceattacksmachine}
R.~Shokri, M.~Stronati, C.~Song, and V.~Shmatikov.
\newblock Membership inference attacks against machine learning models, 2017{\natexlab{b}}.
\newblock URL \url{https://arxiv.org/abs/1610.05820}.

\bibitem[Sohl-Dickstein et~al.(2015)Sohl-Dickstein, Weiss, Maheswaranathan, and Ganguli]{sohldickstein2015deepunsupervisedlearningusing}
J.~Sohl-Dickstein, E.~A. Weiss, N.~Maheswaranathan, and S.~Ganguli.
\newblock Deep unsupervised learning using nonequilibrium thermodynamics, 2015.
\newblock URL \url{https://arxiv.org/abs/1503.03585}.

\bibitem[Student(1908)]{student1908}
Student.
\newblock The probable error of a mean.
\newblock \emph{Biometrika}, 6\penalty0 (1):\penalty0 1--25, 1908.
\newblock \doi{10.1093/biomet/6.1.1}.

\bibitem[Tashiro et~al.(2021)Tashiro, Song, and Ermon]{tashiro2021csdi}
Y.~Tashiro, J.~Song, and S.~Ermon.
\newblock Csdi: Conditional score-based diffusion models for imputation.
\newblock \emph{NeurIPS}, 2021.

\bibitem[{von Loessl}(2023)]{privacy_concerns}
V.~{von Loessl}.
\newblock Smart meter-related data privacy concerns and dynamic electricity tariffs: Evidence from a stated choice experiment.
\newblock \emph{Energy Policy}, 180:\penalty0 113645, 2023.
\newblock ISSN 0301-4215.
\newblock \doi{https://doi.org/10.1016/j.enpol.2023.113645}.
\newblock URL \url{https://www.sciencedirect.com/science/article/pii/S0301421523002306}.

\bibitem[Wilcoxon(1945)]{wilcoxon1945}
F.~Wilcoxon.
\newblock Individual comparisons by ranking methods.
\newblock \emph{Biometrics Bulletin}, 1\penalty0 (6):\penalty0 80--83, 1945.
\newblock \doi{10.2307/3001968}.

\bibitem[Wu et~al.(2025)Wu, Pang, Liu, and Wu]{wu2025winning}
X.~Wu, Y.~Pang, T.~Liu, and S.~Wu.
\newblock Winning the midst challenge: New membership inference attacks on diffusion models for tabular data synthesis.
\newblock \emph{arXiv preprint arXiv:2503.12008}, 2025.

\bibitem[Xu et~al.(2019{\natexlab{a}})Xu, Skoularidou, Cuesta-Infante, and Veeramachaneni]{xu2019modeling}
L.~Xu, M.~Skoularidou, A.~Cuesta-Infante, and K.~Veeramachaneni.
\newblock Modeling tabular data using conditional gan.
\newblock In \emph{Advances in Neural Information Processing Systems}, 2019{\natexlab{a}}.

\bibitem[Xu et~al.(2019{\natexlab{b}})Xu, Skoularidou, Cueto, and Gonz{\'a}lez]{xu2019ctgan}
L.~Xu, M.~Skoularidou, A.~Cueto, and J.~Gonz{\'a}lez.
\newblock Modeling tabular data using conditional gan.
\newblock In \emph{Advances in Neural Information Processing Systems 32}, pages 7335--7345, 2019{\natexlab{b}}.

\bibitem[Yilmaz and Korn(2022)]{synthetic_demand_gan}
B.~Yilmaz and R.~Korn.
\newblock Synthetic demand data generation for individual electricity consumers : Generative adversarial networks (gans).
\newblock \emph{Energy and AI}, 9:\penalty0 100161, 2022.
\newblock ISSN 2666-5468.
\newblock \doi{https://doi.org/10.1016/j.egyai.2022.100161}.
\newblock URL \url{https://www.sciencedirect.com/science/article/pii/S2666546822000209}.

\bibitem[Yoon et~al.(2019)Yoon, Jarrett, and Schaar]{timegan}
J.~Yoon, D.~Jarrett, and M.~Schaar.
\newblock Time-series generative adversarial networks.
\newblock 12 2019.

\end{thebibliography}

\end{document}